%% file: iclr2024_conference.tex
\documentclass{article} 
\usepackage{iclr2024_conference,times}

\input{math_commands.tex}

\usepackage{hyperref}
\usepackage{url}

\title{
Linearizing Models for Efficient yet Robust Private Inference}

\author{Sreetama Sarkar$^{1}$\thanks{Equally contributing authors.}, \ \ Souvik Kundu$^{2\ast}$ \ \ Peter A. Beerel$^{1}$\\
$^{1}$Universiy of Southern California, Los Angeles, USA \ \ \ \
$^{2}$Intel Labs, USA \\  
\tt\small\{sreetama, pabeerel\}@usc.edu \ \ \ \ \tt\small\{souvikk.kundu\}@intel.com
}

\iclrfinalcopy 
\begin{document}

\maketitle

\begin{abstract}
The growing concern about data privacy has led to the development of private inference (PI) frameworks in client-server applications
which protects both data privacy and model IP.  However, the cryptographic primitives required yield significant latency overhead which limits its wide-spread application. At the same time, changing environments demand the PI service to be robust against various naturally occurring and gradient-based perturbations. Despite several works focused on the development of latency-efficient models suitable for PI, the impact of these models on robustness has remained unexplored. Towards this goal, this paper presents RLNet, a class of robust linearized networks that can yield latency improvement via reduction of high-latency ReLU operations while improving the model performance on both clean and corrupted images. In particular, RLNet models provide a ``triple win ticket" of improved classification accuracy on clean, naturally perturbed, and gradient-based perturbed images using a shared-mask shared-weight architecture with over an order of magnitude fewer 
ReLUs than baseline models.
To demonstrate the efficacy of RLNet, we perform extensive experiments with ResNet and WRN models on CIFAR-10, CIFAR-100, and Tiny-ImageNet. Our experimental evaluations show that RLNet can yield models with up to $11.14\times$ fewer ReLUs, with accuracy close to the all-ReLU models, on clean, naturally perturbed, and gradient-based perturbed images. Compared with the SoTA non-robust linearized models at similar ReLU budgets, RLNet achieves an improvement in adversarial accuracy of up to $\mathord{\sim}47\%$, naturally perturbed accuracy up to $\mathord{\sim}16.4\%$, while improving clean image accuracy up to $\mathord{\sim}1.5\%$.
\end{abstract}

\section{Introduction}
\input{01_introduction}


\section{Proposed Approach}
\input{03_approach}

\vspace{-4mm}
\section{Experimental Results}
\input{04_experiments}

\vspace{-2mm}
\section{Conclusions}
The latency overhead for PI in CNNs can be largely attributed to the presence non-linear ReLU units. Latency efficient PI methods have devised ReLU reduction techniques in CNNs. However, robustness of these partially linearized models remain unexplored. In this paper, we propose RLNet, a class of shared-mask shared-weight conditional models that yields close to baseline accuracy against clean, naturally perturbed as well as adversarial images with up to 11.14$\times$ fewer ReLU.
Compared with its SoTA non-robust counterpart, RLNet models improve adversarial accuracy up to $\mathord{\sim}47\%$, naturally perturbed accuracy up to $\mathord{\sim}16.4\%$, while improving clean image accuracy up to $\mathord{\sim}1.5\%$.
Exploring robustness of vision transformer models for latency efficient PI can be an interesting future research in this direction.

\bibliography{iclr2024_conference}
\bibliographystyle{iclr2024_conference}

\appendix
\section{Appendix}
\subsection{Related Work}
\input{02_related_work}
\input{appendix}

\end{document}

%% file: math_commands.tex
\usepackage{amsmath,amsfonts,bm}

\def\eqref#1{equation~\ref{#1}}

\def\1{\bm{1}}

\DeclareMathAlphabet{\mathsfit}{\encodingdefault}{\sfdefault}{m}{sl}
\SetMathAlphabet{\mathsfit}{bold}{\encodingdefault}{\sfdefault}{bx}{n}



%% file: 01_introduction.tex
\label{sec:intro}
In recent years, there has been a growing concern about data privacy, particularly in applications that rely on Machine Learning as a Service (MLaaS) \cite{kundu2021analyzing, ma2021undistillable}
\begin{wrapfigure}{r}{0.55\textwidth}
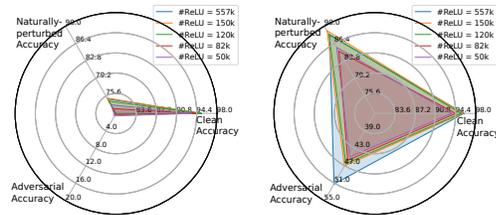

\vspace{-4mm}
  \begin{center}
    \includesvg[width=0.24\textwidth]{images/result2b}
    \includesvg[width=0.24\textwidth]{images/result2a}
  \end{center}
  \vspace{-3mm}
  \caption{Clean, naturally-perturbed, and adversarial accuracy of ResNet18 for non-robust linearized models (left) and RLNet (right) on CIFAR-10 for different ReLU budget. SoTA linearized models lack robustness against natural and adversarial perturbations, while RLNet performs well on all three fronts outperforming its non-robust counterpart even in clean accuracy. (Note: all axes are not on same scale.)}
  \vspace{-8mm}
  \label{fig:intro}
\end{wrapfigure}
in which user data is sent to the cloud to perform inference. This has led to the development of private inference (PI) frameworks, where the server performs 
computations on a client’s encrypted data, preserving both model and data privacy. 
Nevertheless, the implementation of required cryptographic protocols including homomorphic encryption \cite{Reagen2020CheetahOA} and secure multi-party communication \cite{Mishra2020DelphiAC} dramatically increase the computation and communication latency, making it impractical to perform private inference on large scale models. A deeper look into the inference latency of these cryptographic primitives applied to CNNs suggest that the overhead associated with the ReLU layers is
${\sim}340\times$ more than that of the convolutional layers \cite{Kundu2023LearningTL}. This has sparked an interest in the reduction of ReLU non-linearity for latency-efficient private inference of CNNs. 


Techniques for ReLU reduction without compromising accuracy has been extensively studied in the literature \cite{Jha2021DeepReDuceRR, Ghodsi2020CryptoNASPI, Kundu2023LearningTL}.
However, these models lack robustness against naturally or adversarially perturbed images, necessary to ensure trustworthy deployment, as demonstrated in Figure \ref{fig:intro} (left). The natural perturbations may stem from seasonal changes (summer versus winter), environmental conditions (rain, fog, and snow), and/or camera noise and blurring effects \cite{Hendrycks2019BenchmarkingNN}.  Adversarial perturbations~\cite{Goodfellow2014ExplainingAH, Madry2017TowardsDL} are carefully crafted using gradient-based attacks such that they are not noticeable to the human eye but cause misclassifications. These issues are particularly important for applications such as household robots \cite{amazonScienceBehind}, which operates in dynamic environments like changing lighting conditions, and collects enormous amount of user's sensitive personal data, necessitating PI.
To the best of our knowledge, robustness against natural and adversarial perturbations of the partially linearized models is yet to be explored. With this motivation, the objective of this paper is to understand, analyze, and improve the relevant trade-offs 
as well as to build a training framework that guarantees both privacy and robustness. 

The failure of models on naturally perturbed images is caused by a shift in the distribution between the inference and training data.
To improve generalization to these distribution shifts, data augmentation techniques \cite{Hendrycks2019AugMixAS, Wang2021AugMaxAC} are commonly used, but they cannot defend against strong gradient-based perturbations.
Adversarial training \cite{Madry2017TowardsDL, freeat} is the most popular and effective defense against gradient-based adversarial attacks. However, the improved adversarial accuracy does not guarantee improved performance on natural perturbations and is often achieved at the cost of a significant reduction in clean accuracy \cite{Zhang2019TheoreticallyPT, Zi2021RevisitingAR}. 
One means of achieving high clean, naturally perturbed, and adversarial accuracy is to have multiple models in the server and switch models in real time during inference. 
However, this triples off-line processing and storage requirements, induces energy and latency 
overheads when switching between models, and increases the logistics of maintaining 
and delivering multiple models to the customer.
To solve this problem, this paper proposes robust linearized networks termed RLNet, a class of shared-mask shared-weight conditional models, that provides a configurable trade-off between accuracy and robustness while improving latency via reduced ReLU operations. As shown in Figure \ref{fig:intro} (right), RLNet maintains close to baseline accuracies on clean, naturally perturbed, and adversarial images for ReLU reduction  up to $11.14\times$. 

\textbf{Our Contributions:}
Our contributions in this work are three-fold. \textit{Firstly,} we propose a training framework that achieves a \textit{``triple win ticket''}, that is, improved accuracy on clean, naturally perturbed, and adversarial images. Specifically, we implement a conditional learning \cite{Wang2020OnceforAllAT, Kundu2023FLOATFL} strategy using dual Batch Normalization (BN) to build a multi-path model, to retain performance on all three fronts. 
Unlike \cite{Wang2020OnceforAllAT, Kundu2023FLOATFL}, our model requires no additional layers or parameters and, hence, incurs no increase in computational overhead. 
\textit{Secondly,} we develop a fine-tuning framework for partial ReLU (PR) model distillation from an all ReLU (AR) model such that it provides both accuracy and robustness with a reduced PI latency budget. We present RLNet, a class of conditionally trained PR models, efficiently fine-tuned to yield improved performance. \textit{Finally,} we conduct extensive experimental evaluations to demonstrate the efficacy of the proposed training framework across various models on multiple datasets.
Compared with the SoTA non-robust linearized model \cite{Kundu2023LearningTL}, at similar ReLU budgets, RLNet achieves an improved adversarial accuracy of up to $\mathord{\sim}47\%$, naturally perturbed accuracy up to $\mathord{\sim}16.4\%$, while improving clean image accuracy up to $\mathord{\sim}1.5\%$.

%% file: 03_approach.tex
We propose a three-stage pipeline for training RLNet models that consists of training a robust AR teacher, generating a ReLU mask for achieving a target number of ReLU operations in the PR model, and finally, fine-tuning the PR model with the ReLU mask frozen to reduce the performance gap with 
the AR model.
RLNets have two different modes of operation: \textit{normal mode} which targets clean and naturally-perturbed images, and \textit{adversarial mode} which targets adversarial images. The model may be equipped to automatically switch modes based on prediction confidence, as suggested in \cite{stutz2020confidence}. We leverage data augmentation, adversarial distillation, and dual BN to achieve our three-fold objective: clean accuracy (CA), naturally perturbed accuracy (NPA), and adversarial accuracy (AdvA). 
 
\vspace{-2mm}
\subsection{Data Augmentation}
\label{sec:data_aug}
In our framework, we generate augmented images ($x_{aug}$) using Augmix \cite{Hendrycks2019AugMixAS} to improve generalization against natural perturbations. Augmix (augment+mix) involves applying a chain of simple augmentation operations including translation, rotation, shear, auto-contrast, and linearly combining or mixing the augmented images. Augmentations based on brightness, contrast, colour, used in CIFAR-10-C or CIFAR-100-C, which are used for robustness evaluation, are excluded from the set of augmentations during training.
Images from multiple augmentation chains (by default, 3) are then mixed together using a set of convex coefficients randomly sampled from a Dirichlet distribution. 
\vspace{-2mm}
\subsection{Adversarial Training}
\label{sec:adv_train}
We generate adversarial images ($x_{adv}$) using the PGD attack \cite{Madry2017TowardsDL} for robust teacher training. Attacked images are generated by finding perturbations within the maximum perturbation strength $\epsilon$ that maximizes the cross-entropy (CE) loss, following
$x_{adv} = \underset{||x_{adv}-x||_{p} \leq \epsilon}{\arg \max~} 
            CE(\phi_{T}^{\lambda=1}(x_{adv}),y)$. 
The AR teacher model is represented as $\phi_{T}$ and the PR student model is denoted as $\phi_{S}$. $\lambda=0$ denotes the path trained using clean and augmented images and $\lambda=1$ denotes the path trained using adversarial images. During training of the PR model from the AR model through distillation, $x_{adv}$ is generated by maximizing the KL divergence loss, inspired from RSLAD \cite{Zi2021RevisitingAR}, given by $ x_{adv} = \underset{||x_{adv}-x||_{p} \leq \epsilon}{\arg \max~} 
            KL(\phi_{S}^{\lambda=1}(x_{adv}),\phi_{T}^{\lambda=0}(x))$.

\vspace{-4mm}
\subsection{Dual Batch Normalization}
\label{sec:dualBN}
\begin{wrapfigure}{r}{0.4\textwidth}
\vspace{-8mm}
  \begin{center}
    \includegraphics[width=0.35\textwidth]{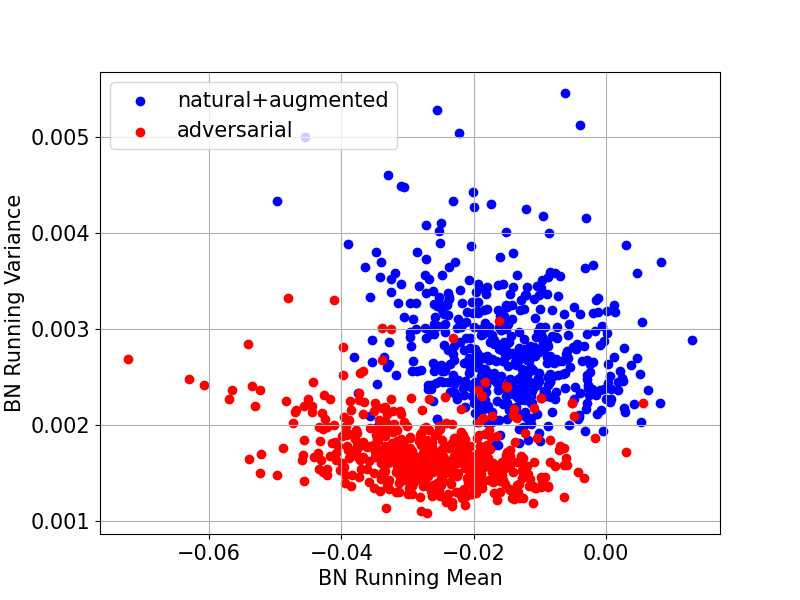}
  \end{center}
  \vspace{-3mm}
  \caption{Running mean and variance of the last BN layer of ResNet18 trained on Tiny-ImageNet using dual BN}
  \vspace{-5mm}
  \label{fig:bn_stats}
\end{wrapfigure}
Previous work \cite{Wang2020OnceforAllAT, Xie2019AdversarialEI, Wang2021AugMaxAC} has pointed out that separation of the BN statistics is critical for a single model to perform well on both clean and adversarial images. BN relies on the fact that input images have the same underlying distribution. Because the distribution of adversarial images is significantly different from that of clean images, adversarial training fails to perform well on clean images. Xie et al. \cite{Xie2019AdversarialEI} demonstrated that adversarial images can in fact improve clean accuracy only if the BN statistics of adversarial images do not interfere with clean training.  

\begin{figure}[htbp]
    \centering
    \vspace{-6mm}
    \includesvg[width=0.7\textwidth]{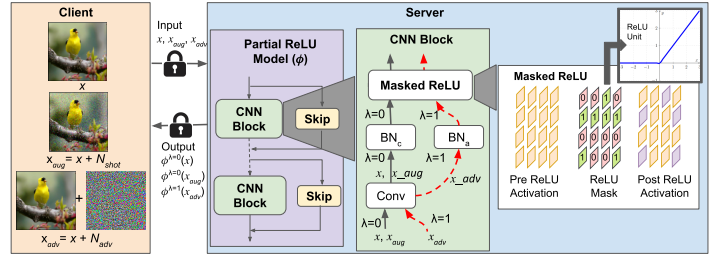}
    \vspace{-12mm}
    \caption{RLNet framework: model architecture and inference path. Here, $\lambda=0$ and $1$ correspond to clean and adversarial paths respectively. We use the same path as clean for classifying against natural perturbations. This means that unless the perturbation is attacker-driven, we use the $\lambda=0$ path for inference.}
    \vspace{-4mm}
    \label{fig:overview}
\end{figure}

In our case, we build a model with two distinct paths for $\lambda=0$ and $\lambda=1$, for two different modes of operation, which differ only in BN layers. $BN_{c}$ denotes the BN layers for clean, augmented, and $BN_{a}$ for adversarial images (Figure \ref{fig:overview}). \cite{Xie2019AdversarialEI} uses a triple BN formulation when training with clean, augmented and adversarial images. We 
do not use separate BN for clean and augmented images, 
because we observe that using simple augmentations actually makes the dataset more diverse and improves clean accuracy rather than adversely affecting it. The BN statistics of a ResNet18 model trained on Tiny-ImageNet, illustrated in Figure \ref{fig:bn_stats}, clearly shows two distinct clusters, justifying this choice. We further demonstrate that a triple BN formulation, separating clean and augmented images, is redundant (see Figure \ref{fig:rlnet_vs_triplebn_individual}(a)).

\subsection{Training Framework}

\subsubsection{Training a Robust Teacher}
The PR model is distilled from a teacher model, which has the same architecture as the PR model but with all ReLUs present. Therefore, the first step is to train a teacher model that is robust against natural as well as adversarial perturbations and still retains its clean accuracy. 
The teacher model is trained on a classification task with input $x$ and corresponding labels $y$. $x_{aug}$ and $x_{adv}$ are generated from $x$ as described in Sections \ref{sec:data_aug} and \ref{sec:adv_train} respectively. 
$\phi_{T}$ is trained by minimizing the loss function $\mathcal{L}_{T}$ with respect to $x$, $x_{aug}$, and $x_{adv}$, as shown in Equation \ref{eq:teacher_loss}, where CE denotes cross-entropy loss.
\begin{align}
\begin{split}
     \mathcal{L}_{T}=(CE(\phi_{T}^{\lambda=0}(x),y)+CE(\phi_{T}^{\lambda=0}(x_{aug}),y)+CE(\phi_{T}^{\lambda=1}(x_{adv}),y))/3 
\end{split}
\label{eq:teacher_loss}
\end{align}

\subsubsection{ReLU Mask Identification}
Given a global ReLU budget, the number of ReLU units per layer has to be determined. \cite{Kundu2023LearningTL} found an inverse relation between pruning sensitivity and ReLU sensitivity and formulated an approach to determine the layerwise ReLU budget. Following \cite{Kundu2023LearningTL}, we allocate the layerwise ReLU count, given by $r_{l}$ for ReLU layer $l$. A binary ReLU mask is initialized in layer $l$, with $r_{l}$ 1's in random locations, where 1's and 0's indicate the presence or absence of ReLU units. ReLU returns 0 if the activation value is negative but retains the original value if the activation value is positive. Hence, the absence of a ReLU unit only makes a difference for negative activation values. 
At the end of each epoch, the ReLU mask is updated with 1's in the top $r_{l}$ locations, where the absolute difference between the post ReLU activation map of the AR and the PR model is the highest. During mask search, the masks and weights of the PR model are updated in parallel. The model weights are updated for each mini-batch using the loss function $\mathcal{L}_{stage2}$ described in Equation \ref{eq:stage2}.  Note, the activation maps for absolute difference calculation and mask updating are obtained using clean images only. 


\subsubsection{Three-way Robust Distillation}
We formulate three different loss functions for our dual BN framework $\mathcal{L}_{ce}$, $\mathcal{L}_{kl}$ and $\mathcal{L}_{PRAM}$ based on CE loss, KL divergence based distillation loss \cite{hinton2015distilling}, and post-ReLU activation mismatch (PRAM) loss \cite{Kundu2023LearningTL}. 
\begin{align}
\scriptsize
     \mathcal{L}_{ce}=& (CE(\phi_{S}^{\lambda=0}(x),y)+CE(\phi_{S}^{\lambda=0}(x_{aug}),y)+ 
     CE(\phi_{S}^{\lambda=1}(x_{adv}),y))/3 \\
     \mathcal{L}_{kl}=& (KL(\phi_{S}^{\lambda=0}(x),\phi_{T}^{\lambda=0}(x))+KL(\phi_{S}^{\lambda=0}(x_{aug}),\phi_{T}^{\lambda=0}(x))+
    KL(\phi_{S}^{\lambda=1}(x_{adv}),\phi_{T}^{\lambda=0}(x)))/3
\label{eq:kl_loss}
\end{align}
Each of these loss functions can have three different components corresponding to three different versions of the input, $x$, $x_{aug}$ and $x_{adv}$, and their corresponding paths through the model. While $\mathcal{L}_{ce}$ ensures learning from the original hard labels $y$, $\mathcal{L}_{kl}$ enforces that the student model learns the output for clean, augmented, and adversarial images through distillation from the clean predictions of the teacher.

PRAM loss \cite{Kundu2023LearningTL} is given by 
\begin{align}
    PRAM(x) = {\left\lVert\frac{\Psi_{PR}^m(x)}{\lVert \Psi_{PR}^m(x)\rVert_2} - \frac{\Psi_{AR}^m(x)}{\lVert \Psi_{AR}^m(x)\rVert_2}\right\lVert_2}
\end{align}
where $\Psi_{PR}^m(x)$ and $\Psi_{AR}^m(x)$ denote the $m^{th}$ pair of post ReLU activation maps for input $x$ for the PR and the AR model. $\mathcal{L}_{PRAM}$ ensures feature similarity between the AR and PR model and is used only during the final fine-tuning to reduce the performance gap. We consider PRAM loss for $x$ and $x_{aug}$, as we find PRAM loss for $x_{adv}$ do not provide significant benefits (see Section \ref{sec:ablation}). 
\begin{align}
    \mathcal{L}_{PRAM}=&0.5\times(PRAM(x)+PRAM(x_{aug}))
    \label{eq:PRAM}
\end{align}
The necessity of each of these loss functions has been justified through ablation studies in Section \ref{sec:ablation}. 
Weight updates for the student model $\phi_{S}$ during mask identification and final fine-tuning are performed through optimizing $\mathcal{L}_{stage2}$ and $\mathcal{L}_{stage3}$ respectively. 
\begin{align}
\scriptsize
\label{eq:stage2}
\mathcal{L}_{stage2}= & \mathcal{L}_{kl}+\mathcal{L}_{ce} \\
\label{eq:stage3}
\mathcal{L}_{stage3}= & \mathcal{L}_{kl}+\mathcal{L}_{ce}+\beta*\mathcal{L}_{PRAM}
\end{align}

%% file: 04_experiments.tex
\subsection{Experimental Setup}
\textbf{Models and Dataset}
We evaluate the efficacy of our proposed robust linearization approach on three different datasets: CIFAR-10, CIFAR-100 \cite{Krizhevsky2009LearningML}, and Tiny-ImageNet \cite{hansen2015tiny} using 3 different models, ResNet-18, ResNet-34 \cite{he2016deep}, and WRN-22-8 \cite{zagoruyko2016wide}. Evaluation against natural perturbations is performed using the common image corruption benchmarks: CIFAR-10-C, CIFAR-100-C, and Tiny-ImageNet-C \cite{Hendrycks2019BenchmarkingNN}, which are generated by adding fifteen different corruptions (like gaussian noise, shot noise, frost, snow, brightness, contrast) at five severity levels to the original dataset.

\textbf{Evaluation Metrics}
Performance on clean, naturally-perturbed, and adversarial images is evaluated using CA, NPA, and AdvA respectively. NPA is the average accuracy over all 15 corruptions, where accuracy on each corruption is averaged over five severity levels. In addition, we also report mean corruption error or $mCE$ on naturally perturbed samples. $mCE$ on CIFAR-10-C and CIFAR-100-C is evaluated as $1-NPA$, whereas
for Tiny-ImageNet the error for each type of corruption is first normalized by the corruption error of a baseline model and then averaged \cite{Hendrycks2019BenchmarkingNN, Wang2021AugMaxAC}. 
CA and AdvA are given by the Top-1 accuracies evaluated on the original dataset and using PGD-7 attack \cite{Madry2017TowardsDL}, if not mentioned otherwise.
Communication savings is the ratio of communication cost of an AR model to that of a PR model. 

\begin{table*}[h]
    \small\addtolength{\tabcolsep}{-2pt}
    \begin{center}
    \resizebox{\textwidth}{!}{
        \begin{tabular}{l|l|c|cccc|cccc|c|c}
        \toprule
 		\textbf{Dataset}  & \textbf{Model} & \textbf{State} & \multicolumn{4}{c|}{\textbf{RLNet}} & \multicolumn{4}{c|}{\textbf{SeNet \cite{Kundu2023LearningTL}}}  & \textbf{\#ReLU} & \textbf{Comm.}\\
 		 & & & \textbf{CA(\%)} & \textbf{NPA(\%)} & \textbf{mCE}$\downarrow$ &  \textbf{AdvA(\%)} & \textbf{CA(\%)} & \textbf{NPA(\%)} & \textbf{mCE}$\downarrow$ &  \textbf{AdvA(\%)} & \textbf{(k)} & \textbf{Savings}\\
 		\midrule
		\multirow{4}{*}{CIFAR-10}  & \multirow{4}{*}{ResNet18} & AR & \textbf{95.88} & \textbf{88.15} & \textbf{0.12} & \textbf{51.18}  & 95.23 & 74.98 & 0.25 & 0.01 & 557 & 1$\times$\\
		                        &    & PR    & \textbf{93.71}   & \textbf{85.62}  & \textbf{0.14} & \textbf{44.95}
                          & 93.41 & 73.78 & 0.26 & 0.58 & 50 & 11.14$\times$    \\
		                        &    & PR    & \textbf{94.43}  & \textbf{84.83} & \textbf{0.15} & \textbf{45.75}  & 94.24 & 72.99 & 0.27 & 0.28 & 82 & 6.79$\times$      \\ 
                          	 &    & PR    & \textbf{95.36}	& \textbf{88.35}	& \textbf{0.12} & \textbf{46.52}  &  95.12 & 74.53 & 0.25 & 0.13 & 120 & 4.64$\times$      \\ 
		                        &    & PR    & \textbf{95.67}   & \textbf{89.04} & \textbf{0.11} & \textbf{47.04}  & 95.0 & 75.17 & 0.25 & 0.12 & 150 & 3.71$\times$      \\ 
		\midrule
            \multirow{9}{*}{CIFAR-100}  & \multirow{3}{*}{ResNet18} & AR & \textbf{78.68} & \textbf{63.31} & \textbf{0.37} & \textbf{26.99}  & 77.59 & 48.67 & 0.51 & 0.04 & 557 & 1$\times$ \\ 
		                        &    & PR    & \textbf{74.87} & \textbf{60.72} & \textbf{0.39} & \textbf{22.07} & 74.51 & 48.71 & 0.51 & 0.07 & 50 & 11.14$\times$ \\ 
		                        &    & PR    & \textbf{77.00} & \textbf{64.45} & \textbf{0.36} & \textbf{22.99} & 76.67 & 49.90 & 0.50 & 0.09 & 100 & 5.57$\times$ \\ 
            \cmidrule{2-13}
                                        & \multirow{3}{*}{ResNet34} & AR & \textbf{79.71} & \textbf{65.25} & \textbf{0.35} & \textbf{26.57} & 78.25 & 49.63 & 0.50 & 0.29 & 967 & 1$\times$\\
		                        &    & PR    & \textbf{73.72} & \textbf{61.27} & \textbf{0.39} & \textbf{22.03} & 72.84 & 51.67 & 0.48 & 0.30 & 80 & 12$\times$ \\
		                        &    & PR    & 75.76   & \textbf{63.90} & \textbf{0.36} & \textbf{23.17}  & \textbf{76.07} & 51.59 & 0.48 & 0.26 & 200 & 4.8$\times$ \\		
            \cmidrule{2-13}
                                        & \multirow{3}{*}{WRN22-8} & AR & \textbf{80.96} & \textbf{66.38} & \textbf{0.34} & \textbf{29.69} & 79.77 & 49.62 & 0.50 & 0.03 & 1393 & 1$\times$ \\ 
		                        &    & PR  & \textbf{80.53} & \textbf{66.58} & \textbf{0.33} & \textbf{25.85} & 79.75 & 50.73 & 0.49 & 0.04 & 240 & 5.8$\times$ \\ 
		                        &    & PR    & \textbf{80.66} & \textbf{67.49} & \textbf{0.33} & \textbf{25.61} & 79.95 & 51.11 & 0.49 & 0.05 & 300 & 4.64$\times$ \\
            \midrule
            \multirow{3}{*}{Tiny-ImageNet}  & \multirow{3}{*}{ResNet18} & AR & \textbf{67.22} & \textbf{37.99} & \textbf{0.81} & \textbf{20.52} & 66.1 & 26.91 & 0.96 & 0.08 & 2228 & 1$\times$ \\
	& & PR & 58.72 & \textbf{31.31} & \textbf{0.90} & \textbf{12.85} &  \textbf{59.31} & 24.01 & 0.99 & 0.15 & 150 & 14.85$\times$ \\
	  & & PR & \textbf{66.57} & \textbf{39.45} & \textbf{0.79} & \textbf{15.99} &  66.18 & 27.54 & 0.95 & 0.19 & 300 & 7.43$\times$ \\
        \bottomrule
    	\end{tabular}
     }
    \caption{Performance evaluation of RLNet on CIFAR-10, CIFAR-100, Tiny-ImageNet and comparison with SeNet \cite{Kundu2023LearningTL}. The higher accuracy and lower mCEs are highlighted in bold.}
     \label{tab:results1}
  \end{center}
\end{table*}
\vspace{-4mm}
\subsection{Results and Analysis}
\begin{wrapfigure}{r}{0.35\textwidth}
  \begin{center}
    \includegraphics[width=0.3\textwidth]{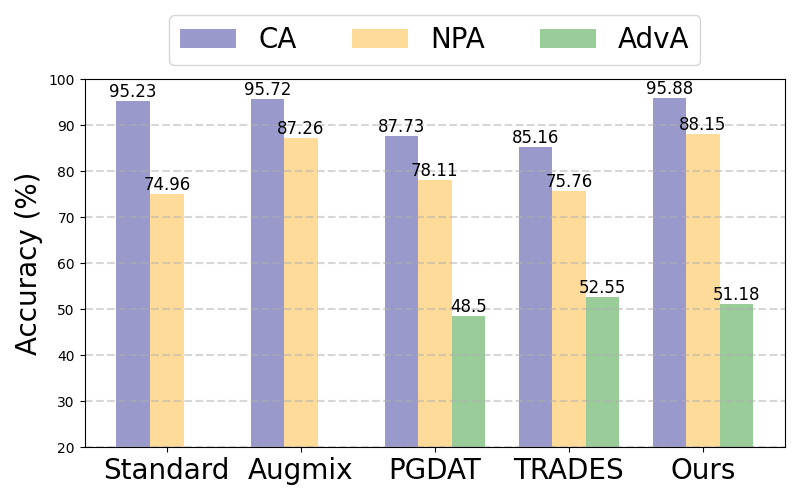}
    \caption{CA, NPA, and AdvA for ResNet18 on CIFAR-10 for different training modes}
    \label{fig:robust_teacher}  \end{center}
   \vspace{-5mm}
\end{wrapfigure}

\textbf{AR Model} In Figure \ref{fig:robust_teacher}, we compare our conditional teacher training approach with individual SoTA training methods using natural, augmented, and adversarial images.
Our approach achieves improvement in CA by 0.65\% over a standard trained model, NPA by 0.89\% over a model trained using Augmix \cite{Hendrycks2019AugMixAS}, and AdvA by 2.68\% over a model trained using PGDAT \cite{Madry2017TowardsDL} for ResNet18 on CIFAR-10. While TRADES \cite{Zhang2019TheoreticallyPT} achieves the best AdvA among these training methods, it comes at the cost of a degradation in CA by 10.72\%.

\textbf{PR Model}
The performance of RLNet on CIFAR-10, CIFAR-100, and Tiny-Imagenet for different ReLU budget is presented in Table \ref{tab:results1}. On CIFAR-10, our method achieves $3.7\times$ ReLU reduction with almost no degradation in CA and NPA and AdvA degradation of 4.14\%. We achieve up to $5.8\times$ ReLU reduction on CIFAR-100 dataset with a nominal reduction in CA of 0.4\%, AdvA of 3.84\%, and with no degradation in NPA. In fact we observe that some amount of ReLU reduction actually boosts NPA. NPA for ResNet18 with 150k ReLU on CIFAR-10-C, 100k ReLU on CIFAR-100-C, and 300k ReLU on Tiny-ImageNet-C are higher than the baseline AR models. For more aggressive ReLU pruning up to $11.14\times$, our model suffers a degradation in CA of only $2.17\%$, NPA of $2.52\%$, and AdvA of $6.23\%$ on CIFAR-10. For the same ReLU budget on CIFAR-100, degradation in CA, NPA, and AdvA are $3.81\%$, $2.59\%$, and $4.92\%$ respectively. For Tiny-Imagenet, we achieve 7.43$\times$ ReLU reduction with close to baseline CA, NPA and AdvA reduction of 4.5\%. We try to ensure minimum degradation in CA through our choice of hyperparameters and loss function, as discussed in Section \ref{sec:ablation}. This explains the higher degradation in AdvA compared to CA and NPA.  

We also compare RLNet models with a SOTA linearized network SeNET \cite{Kundu2023LearningTL}. RLNet consistently outperforms SeNet not only in NPA and AdvA but also in CA. RLNet achieves an improvement in CA over SeNet by $0.67\%$ and $0.3\%$ for ResNet18 on CIFAR-10 for ReLU reduction of $3.7\times$ and $11.14\times$ respectively, $0.78\%$ for WRN22-8 on CIFAR-100 and $0.39\%$ for ResNet18 on Tiny-Imagenet, for a ReLU reduction of $5.8\times$ and $7.43\times$ respectively. RLNet yields up to 16.38\% improvement in NPA and 47\% improvement in AdvA over SeNet models.

\begin{figure}[htbp]
  \centering
  \begin{minipage}{0.3\linewidth}
    \includegraphics[width=\linewidth]{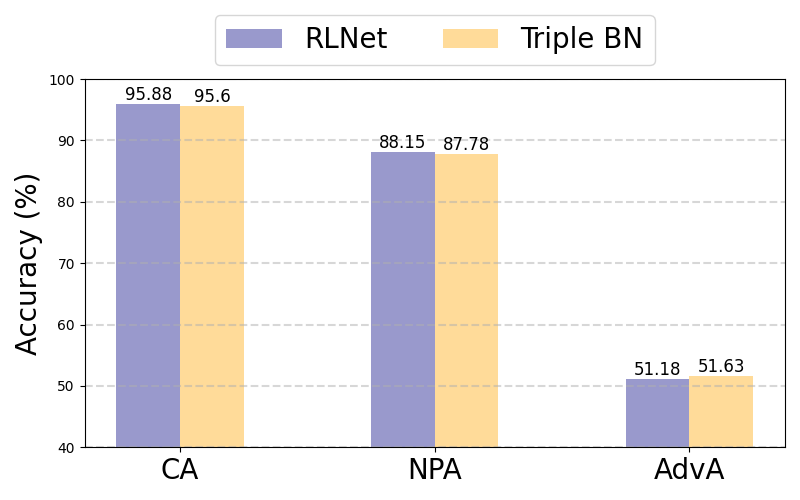}
  \end{minipage}
  \begin{minipage}{0.3\linewidth}
    \includegraphics[width=\linewidth]{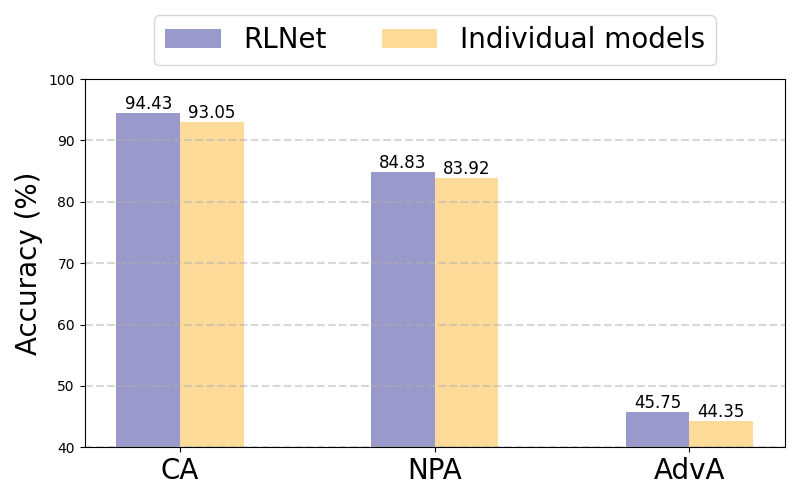}
  \end{minipage}
  \vspace{-3mm}
  \caption{(a) Dual vs triple BN for ResNet18 on CIFAR-10; (b) CA, NPA, and AdvA of RLNet vs separate ResNet18 PR models (ReLU count=82k) trained using standard, Augmix \cite{Hendrycks2019AugMixAS} and PGDAT \cite{Madry2017TowardsDL}.}
    \label{fig:rlnet_vs_triplebn_individual}
  \vspace{-3mm}
\end{figure}

\paragraph{Sufficiency Test of Dual BN}
We demonstrate results using a triple BN formulation, using three separate BN for clean, augmented and adversarial images and compare it with dual BN in Figure \ref{fig:rlnet_vs_triplebn_individual}(a). We observe that triple BN yields no extra benefits, as hypothesized in Section \ref{sec:dualBN}

\paragraph{Sufficiency Analysis of Single ReLU Mask}
The performance of RLNet models is compared with three separate models, trained using standard, Augmix \cite{Hendrycks2019AugMixAS} and PGDAT \cite{Madry2017TowardsDL} training, for the same ReLU count, in Figure \ref{fig:rlnet_vs_triplebn_individual}(b). RLNet achieves improved accuracy over separately-trained individual models in all three cases. This confirms that a shared mask shared weight dual BN model is sufficient to achieve improved performance on all three fronts.


%% file: 02_related_work.tex
\subsubsection{Model Robustness:}
\textbf{Robustness against Natural Perturbations}
Data augmentation techniques \cite{Hendrycks2019AugMixAS, Wang2021AugMaxAC} are commonly used to improve generalization to distribution shifts because they are easy to implement, have low computational overhead, and often also improve clean accuracy.
A number of simple augmentation techniques, including Cutout \cite{Cutout}, occluding random portions in an image,  CutMix \cite{Yun2019CutMixRS}, replacing sections of an image with another image, and MixUp \cite{Zhang2017mixupBE}, generating an image using a linear combination of two different images, have yielded promising results. 
Augmix \cite{Hendrycks2019AugMixAS}, which randomly samples a set of augmentations and linearly combines the augmented images, has shown to be one of the most effective against the common image corruptions benchmarks like CIFAR-10-C and ImageNet-C \cite{Hendrycks2019BenchmarkingNN}. DeepAugment \cite{DeepAugment} generates augmented images by distorting the weights of image-to-image models.
Augmax \cite{Wang2021AugMaxAC} tries to further improve Augmix by learning the mixing coefficients of augmented images to generate harder samples. In this work, we generate augmented images to train our conditional model inspired from the strategy proposed in Augmix \cite{Hendrycks2019AugMixAS}. 

\textbf{Adversarial Robustness}
Adversarial images may be viewed as augmented images, with $l_{p}$-norm bounded perturbations, generated using strong gradient-based attacks.
Data augmentation techniques like AdversarialAugment,\cite{Calian2021DefendingAI}, an enhancement of DeepAugment \cite{DeepAugment}, are designed to achieve better robustness against natural as well as adversarial perturbations.
Training with adversarial images or adversarial training (AT) \cite{Goodfellow2014ExplainingAH, Madry2017TowardsDL} is the most commonly used defense against adversarial attacks, although it causes a degradation in clean accuracy. The increased training overhead of standard AT techniques like PGDAT \cite{Madry2017TowardsDL} has led to the development of compute-efficient alternatives like FreeAT \cite{freeat} and FastAT \cite{wong2020fast}.
Recently, robust distillation methods \cite{Zhang2019TheoreticallyPT, Goldblum2019AdversariallyRD, Zi2021RevisitingAR} have proved to achieve better adversarial robustness compared to standard AT. While TRADES \cite{Zhang2019TheoreticallyPT} uses a self-distillation technique, which utilizes clean predictions of the same model to learn its adversarial predictions, ARD \cite{Goldblum2019AdversariallyRD} and RSLAD \cite{Zi2021RevisitingAR} leverage knowledge distillation to learn from an adversarially robust teacher. We perform robust distillation inspired from RSLAD, to generate robust and latency-efficient PR models from robust AR models.

\subsubsection{Efficient Private Inference}

PI frameworks use cryptograhic protocols such as Homomorphic Encryption (HE)~\cite{juvekar2018gazelle, srinivasan2019delphi} and Additive Secret Sharing (ASS)~\cite{liu2017oblivious, Ghodsi2020CryptoNASPI}. 
Fully HE based protocols like CryptoNets~\cite{gilad2016cryptonets}, CHET~\cite{dathathri2019chet}, TAPAS~\cite{sanyal2018tapas} incur huge computation and communication latency, limiting their applications to networks that are only a few layers deep. 
Gazelle \cite{juvekar2018gazelle}, Delphi~\cite{srinivasan2019delphi}, Cheetah~\cite{reagen2021cheetah} uses HE for linear operations such as convolution and fully-connected layers, while Garbled Circuits (GC)~\cite{yao1986generate} are used for non-linear ReLU operations on the client's encrypted data. 
Delphi~\cite{srinivasan2019delphi} introduces an online-offline topology, where the 
client-data-independent components are pre-computed during an offline phase, enabling online plaintext computation for linear operations. Although this reduces online latency due to linear operations, the input dependent ReLU computation using GC still causes high inference latency. This necessitates either removing ReLU operations~\cite{Cho2022SelectiveNL, Kundu2023LearningTL} or replacing them with some other compute-efficient alternatives like polynomial or quadratic functions~\cite{gilad2016cryptonets, liu2017oblivious, srinivasan2019delphi}. Various approaches for ReLU reduction have been proposed in literature. They range from manually dropping ReLU layers from existing models \cite{Jha2021DeepReDuceRR}, $l_{1}$-regularization based approaches~\cite{Cho2022SelectiveNL} to evolutionary neural architecture search (NAS) techniques~\cite{Ghodsi2020CryptoNASPI} for ReLU reduction. Kundu et al.~\cite{Kundu2023LearningTL, Kundu_2023_CVPR} proposed a 3-stage training approach that meets a target ReLU budget for negligible accuracy reduction. In this paper, we enhance the training pipeline proposed in~\cite{Kundu2023LearningTL} for robust generalization.

\subsubsection{Conditional Learning}
Conditional learning involves training a single model with multiple paths that can be selectively enabled during inference. Conditional models have been used to provide an in-situ trade-off between efficiency and accuracy \cite{Yu2018SlimmableNN} or accuracy and adversarial robustness \cite{Wang2020OnceforAllAT, Kundu2023FLOATFL}. OAT~\cite{Wang2020OnceforAllAT} uses a parameter lambda to control the trade-off between clean and adversarial accuracy through feature-wise linear modulation (FiLM \cite{Perez2017FiLMVR}) layers, conditioned on lambda, and dual BN \cite{Xie2019AdversarialEI}. To remove the FiLM latency overhead, recently, a few works have proposed \cite{Kundu2023FLOATFL, kundu2023sparse} weight-conditioned learning for accuracy robustness trade-off.
We propose a conditional model using dual BN for accuracy robustness trade-off in PR models with no extra parameters or computational overhead. 

%% file: appendix.tex
\subsection{Experimental Results}
\subsubsection{Training Hyperparameters}
The baseline AR model training and final fine-tuning are performed for 240 and 120 epochs on CIFAR and Tiny-ImageNet datasets, using SGD optimizer, with a starting learning rate (lr) of 0.05 for baseline training and 0.01 for fine-tuning. A step lr decay policy is followed during both, where lr drops by a factor of 0.1 at 62.5\%, 75\%, and 87.5\% of the total training epochs. Mask search is performed for 150 and 100 epochs on CIFAR and Tiny-ImageNet, without any drop in learning rate. Distillation temperature is maintained at 4.0, unless otherwise stated. $\beta=1000$ is used in Equation \ref{eq:stage3}. Data augmentations are generated using the default parameters as in \cite{Hendrycks2019AugMixAS}. Adversarial augmentations during training are generated using PGD-7, where the attack is performed for 7 iterations with maximum perturbation strength $\epsilon=8/255$ and step size $\alpha=2/255$. Attack evaluations are also performed using FGSM with $\epsilon=8/255$, PGD-20, with identical parameters as PGD-7 with 20 steps, and Auto-PGD \cite{croce2020reliable}, a variant of AutoAttack, with $\epsilon=8/255$ and 100 iterations.

\subsubsection{Ablation Studies}
\label{sec:ablation}
\textbf{Study of Robust Distillation Loss}
Adversarial robust distillation techniques \cite{Goldblum2019AdversariallyRD, Zi2021RevisitingAR} demonstrate that learning from teacher predictions significantly improves adversarial robustness as compared to regular AT using hard labels. The increase in adversarial robustness is accompanied with a drop in clean accuracy. We try to build a robust distillation approach for our dual BN framework that prioritizes CA, and at the same time achieves a reasonable trade-off between CA, NPA, and AdvA.
\begin{wraptable}{r}{0.55\textwidth}
    \begin{center}
    \vspace{-3mm}
    \small\addtolength{\tabcolsep}{-3.5pt}
        \begin{tabular}{c|ccc|c|c}
        \toprule
 		\textbf{State} & \multicolumn{3}{c|}{\textbf{Accuracy}}   & \textbf{\#ReLU} & \textbf{Loss}\\
 		&\textbf{CA(\%)} & \textbf{NPA(\%)}            &  \textbf{AdvA(\%)} & \textbf{(k)}\\
 		\midrule
		AR & 78.68   & 63.31  & 26.99   & 557 & -\\
		PR    & 70.21   & 58.08  & 23.91  & 50 &  Train$_{CE}$    \\
		PR    & 70.53   & 58.75  & 22.21  & 50 & Train$_{CEKL}$       \\ 
		PR    & 73.47   & 61.13  & 21.82  & 50 & Train$_{KL}$         \\ 
        \bottomrule
    	\end{tabular}
    \caption{Study of robust distillation techniques for ResNet18 PR models on CIFAR-100}
    \vspace{-4mm}
        \label{tab:rd_comparison}
  \end{center}
\end{wraptable}
We formulate three different robust training techniques, inspired from PGD AT \cite{Madry2017TowardsDL}, ARD \cite{Goldblum2019AdversariallyRD}, and RSLAD \cite{Zi2021RevisitingAR}. Train$_{CE}$ uses CE loss for $x$, $x_{aug}$, and $x_{adv}$, similar to PGD AT \cite{Madry2017TowardsDL} for $x$ and $x_{adv}$, which requires no teacher model. Train$_{CEKL}$ uses distillation only for $x_{adv}$, following ARD \cite{Goldblum2019AdversariallyRD}, and CE loss for $x$ and $x_{aug}$. Train$_{KL}$, inspired from RSLAD \cite{Zi2021RevisitingAR}, uses distillation for all three inputs $x$, $x_{aug}$, and $x_{adv}$, where the model learns from the clean predictions of the teacher. The loss functions for these robust training methods are presented in Table \ref{tab:advloss}. 

\begin{table}[!h]
    \small\addtolength{\tabcolsep}{-2.5pt}
    \centering
    \hspace*{-0.5cm}
    \begin{tabular}{c|c}
    \toprule
    \textbf{Method} & \textbf{Loss} \\
    \midrule
        Train$_{CE}$ & $(CE(S^{\lambda=0}(x),y)+CE(S^{\lambda=0}(x^{aug}),y)+CE(S^{\lambda=1}(x^{adv}),y))/3$ \\
    \midrule
        Train$_{CEKL}$ & $(CE(S^{\lambda=0}(x),y)+CE(S^{\lambda=0}(x^{aug}),y)+KL(S^{\lambda=1}(x^{adv}),T^{\lambda=0}(x)))/3$ \\
    \midrule
        Train$_{KL}$ & $(KL(S^{\lambda=0}(x),T^{\lambda=0}(x))+KL(S^{\lambda=0}(x^{aug}),T^{\lambda=0}(x))+KL(S^{\lambda=1}(x^{adv}),T^{\lambda=0}(x)))/3$ \\
    \bottomrule
    \end{tabular}
    \caption{Robust distillation losses for conditional model training}
    \label{tab:advloss}
\end{table}
\begin{wraptable}{r}{0.55\textwidth}
    \begin{center}
        \vspace{-5mm}
    \small\addtolength{\tabcolsep}{-5.3pt}
        \begin{tabular}{ccc|c|c}
        \toprule
 		\multicolumn{3}{c|}{\textbf{Accuracy}} & \textbf{Loss} & \textbf{Temperature}\\
 		\textbf{CA(\%)} & \textbf{NPA(\%)}   &  \textbf{AdvA(\%)} \\
 		\midrule
		73.47	& 61.13	& 21.82   & $\mathcal{L}_{kl}$ & 4.0 \\
		71.08	& 58.6	& 23.32  & $\mathcal{L}_{kl}$ & 1.0 \\
		74.16	& 61.19	& 21.81  & $\mathcal{L}_{kl}+\mathcal{L}_{PRAM}$ & 4.0   \\  
		73.5	& 59.11	& 24.55  & $\mathcal{L}_{kl}+\mathcal{L}_{PRAM}$ & 1.0       \\ 
        \bottomrule
    	\end{tabular}
    \caption{Study of distillation temperature for ResNet18 PR models (ReLU count = 50k) on CIFAR-100}  
    \vspace{-4mm}
    \label{tab:distill_temp}
  \end{center}
\end{wraptable}

Table \ref{tab:rd_comparison} presents the evaluation results for our PR model, trained using these robust training techniques. We observe that KL divergence loss for $x$ and $x_{aug}$ improves CA and NPA by $\sim$3\% compared with CE loss. Hence, we formulate the loss $\mathcal{L}_{kl}$ (Equation \ref{eq:kl_loss}) according to Train$_{KL}$. The degradation in AdvA for Train$_{KL}$ may be attributed to the choice of distillation temperature, favourable towards CA and NPA, as discussed in the next section.

\textbf{Choice of Distillation Temperature}
The choice of distillation temperature plays a critical role in the trade-off between CA and AdvA. In line with RSLAD \cite{Zi2021RevisitingAR}, we observe that using a distillation temperature of 1.0 results in improved AdvA. However, this reduces CA, both with and without PRAM loss, as observed in Table \ref{tab:distill_temp}. Since we prioritize CA, we choose a distillation temperature of 4.0.

\begin{wraptable}{r}{0.55\textwidth}
    \small\addtolength{\tabcolsep}{-3.5pt}
    \begin{center}
    \vspace{0mm}
        \begin{tabular}{c|c|c|c|c|ccc}
        \toprule
 		\textbf{$\mathcal{L}_{kl}$} & \textbf{$\mathcal{L}_{ce}$} & \textbf{$PRAM$} & \textbf{$PRAM$} & \textbf{$PRAM$} & \multicolumn{3}{c}{\textbf{Accuracy}}  \\
 	&& \textbf{$(x)$} & \textbf{$(x_{aug})$} & \textbf{$(x_{adv})$} &	             \textbf{CA} & \textbf{RA} &  \textbf{AdvA} \\
    & & & & & \textbf{(\%)} & \textbf{(\%)} & \textbf{(\%)} \\
    \midrule
         \cmark & \xmark & \xmark  & \xmark & \xmark & 73.47	& 61.13 & 21.82  \\ 
         \cmark & \xmark & \cmark  & \xmark & \xmark & 73.85	& 60.48	& 21.73  \\ 
         \cmark & \xmark & \cmark  & \cmark & \xmark & 74.16	& 61.19	& 21.81  \\ 
         \cmark & \xmark & \cmark  & \cmark & \cmark & 74.20	& 61.53 & 22.19      \\ 
         \cmark & \cmark & \cmark  & \cmark & \xmark & 74.87	& 60.72	& 22.07      \\ 
         \cmark & \cmark & \cmark  & \cmark & \cmark & 74.28	& 61.12	& 21.95      \\ 
    
        \bottomrule
    \end{tabular}
    \caption{PRAM ablation for ResNet18 PR models (ReLU count = 50k) on CIFAR-100}
    \vspace{-4mm}
    \label{tab:attn_ablation}
  \end{center}
\end{wraptable}

\textbf{Study of Robust Feature Similarity Loss}
\label{sec:PRAM_loss}
In this section, we explore the effectiveness of PRAM loss \cite{Kundu2023LearningTL} in our robust training setup. 
PRAM($x$), PRAM($x_{aug}$), and PRAM($x_{adv}$) are the PRAM losses with original, augmented, and adversarial images as input. 
In Table \ref{tab:attn_ablation}, we perform an ablation to understand the contribution of each of these PRAM loss terms. 
We observe that the presence or absence of PRAM loss causes a variation in adversarial accuracy by less than 0.5\%. PRAM($x$), PRAM($x_{aug}$), and PRAM($x_{adv}$) boosts CA, NPA, as well as AdvA, in the absence of $\mathcal{L}_{ce}$.
However, when we incorporate $\mathcal{L}_{ce}$, PRAM($x_{adv}$) is found to degrade both CA and AdvA. Hence, we only use PRAM($x$) and PRAM($x_{aug}$) for feature similarity during fine-tuning (Equation \ref{eq:PRAM}).

\begin{wraptable}{r}{0.57\textwidth}
    \small\addtolength{\tabcolsep}{-4.8pt}
    \begin{center}
          \vspace{-4mm}
        \begin{tabular}{l|c|ccc|c|c}
        \toprule
 		\textbf{Model/}  &  \textbf{State} & \multicolumn{3}{c|}{\textbf{Accuracy}}   & \textbf{\#ReLU} & \textbf{$\mathcal{L}_{ce}$}\\
 		\textbf{Dataset} & &\textbf{CA(\%)} & \textbf{NPA(\%)}            &  \textbf{AdvA(\%)} & \textbf{(k)}\\
 		\midrule
		\multirow{3}{*}{\rotatebox{0}{\makecell{ResNet18 \\ CIFAR-10}}}  & AR & 95.88 & 88.15  & 51.18  & 557 & -\\
		                                & PR    & 93.71   & 85.62  & 44.95  & 50 &  \xmark    \\
		                                & PR    & 93.70   & 85.13  & 43.57  & 50 & \cmark        \\ 
		\midrule
		\multirow{3}{*}{\rotatebox{0}{\makecell{ResNet18 \\ CIFAR-100}}}  & AR & 78.68 & 63.31  & 26.99 & 557 & -\\
		                                & PR    & 74.16   & 61.19  & 21.81  & 50 &  \xmark    \\
		                                & PR    & 74.87   & 60.72  & 22.07  & 50 & \cmark        \\ 
		\midrule                                  
            \multirow{3}{*}{\rotatebox{0}{\makecell{ResNet34 \\ CIFAR-100}}}  & AR & 79.71 & 65.25  & 26.57  & 967 & -\\
		                                & PR    & 73.30   & 61.23  & 21.69  & 80 &  \xmark    \\
		                                & PR    & 73.72   & 61.27  & 22.03  & 80 & \cmark        \\         
        \bottomrule
    	\end{tabular}
     	\caption{Fine-tuning PR model with and without CE loss}
      \vspace{-4mm}
          \label{tab:ce_loss}
  \end{center}
\end{wraptable}
\textbf{Necessity Analysis of CE Loss}
In this section, we analyze whether $\mathcal{L}_{ce}$ helps or hurts performance when used together with $\mathcal{L}_{kl}$ and $\mathcal{L}_{PRAM}$.
In Table \ref{tab:ce_loss}, we present results with and without using $\mathcal{L}_{ce}$ on a number of models and datasets. We observe that $\mathcal{L}_{ce}$ improves CA as well as AdvA for both ResNet18 and ResNet34 models on CIFAR-100, whereas it retains the same CA for ResNet18 on CIFAR-10 dataset. Since our main focus is to minimize degradation in CA, we incorporate $\mathcal{L}_{ce}$ in our training framework. 

\textbf{Performance on Other Attacks}
\begin{figure}[htbp]
  \centering
  \includegraphics[width=\linewidth]{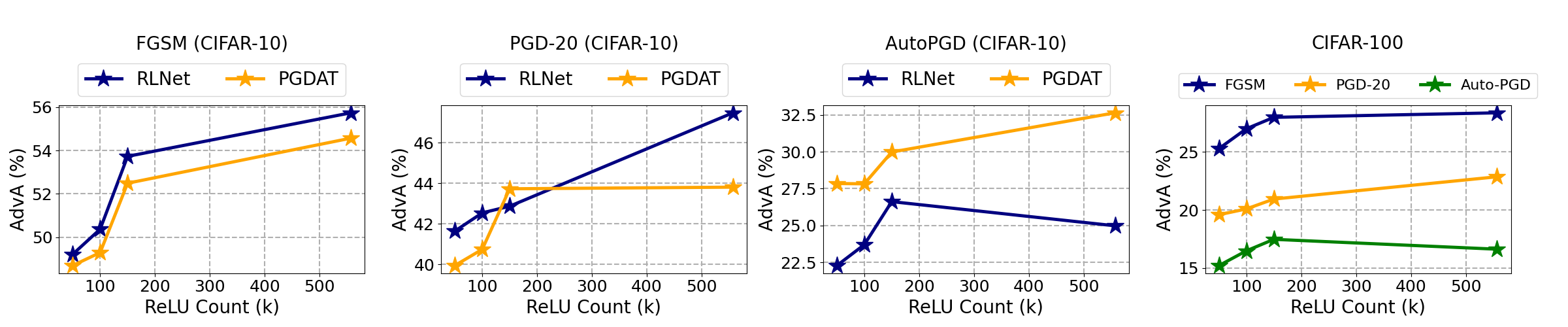}
  \caption{Performance comparison of RLNet vs ResNet-18 models, distilled from a PGD trained teacher, for different ReLU budgets, on (a) FGSM \cite{Goodfellow2014ExplainingAH}, (b) PGD-20\cite{Madry2017TowardsDL} and (c) Auto-PGD \cite{croce2020reliable} attacked images on CIFAR-10 dataset. (d) Performance of ResNet18 for different ReLU budget on FGSM, PGD-20, and AutoPGD attack generated CIFAR-100 test images}
  \vspace{-5mm}
  \label{fig:attacks_eval}
\end{figure}

We evaluate RLNet models against existing SOTA white-box attacks FGSM \cite{Goodfellow2014ExplainingAH}, PGD-20 \cite{Madry2017TowardsDL} and Auto-PGD, a variant of AutoAttack \cite{croce2020reliable}. For ResNet18 on CIFAR-10, we train a robust teacher using PGDAT \cite{Madry2017TowardsDL} and distill PR models for different ReLU budgets using RSLAD \cite{Zi2021RevisitingAR}. In Figure \ref{fig:attacks_eval}(a, b, c), we compare these PR models, distilled from a PGD trained teacher, and focused solely on improving AdvA, to our RLNet models for different ReLU budgets. We still outperform PGDAT PR models for all ReLU counts on FGSM attacked images and most ReLU count on PGD-20 generated images. PGDAT models demonstrate higher robustness against Auto-PGD on CIFAR-10. Attack evaluation performance for RLNet models on CIFAR-100 are presented in Figure \ref{fig:attacks_eval}(d). 
